\documentclass[letterpaper, 10 pt, conference]{ieeeconf}  %

\IEEEoverridecommandlockouts                              
\usepackage{times}
\usepackage{multicol}
\usepackage{graphicx}
\usepackage{subfigure}
\usepackage{epsfig}
\usepackage{amsmath,amssymb,bm}
\usepackage{algorithmic}
\usepackage{url}
\usepackage{color}
\usepackage[ruled]{algorithm2e}
\usepackage{ntheorem}
\usepackage{float}
\usepackage{physics}
\usepackage{cite}
\usepackage{todonotes}
\usepackage{mathtools}
\DeclarePairedDelimiter{\ceil}{\lceil}{\rceil}
\usepackage{ulem}

\newtheorem{problem}{Problem}

\newcommand{\UPDT}[1]{{{#1}}}

\renewcommand{\sout}[1]{\unskip}

\title{\LARGE \bf
Risk-Aware Planning and Assignment for Ground Vehicles using Uncertain Perception from Aerial Vehicles
}
\author{Vishnu D. Sharma$^{\dagger}$, Maymoonah Toubeh$^{\dagger}$, Lifeng Zhou$^{\dagger}$, Pratap Tokekar 
\thanks{M. Toubeh, L. Zhou are with the Department of Electrical \& Computer Engineering, Virginia Tech, Blacksburg, VA 24061, USA.\texttt{\small \{may93, lfzhou\}@vt.edu}. V.D. Sharma and P. Tokekar were with Virginia Tech during completion of this work and are now with the Department of
Computer Science, University of Maryland, College Park, MD 20742, USA (email: {\tt\small \{vishnuds,tokekar\}@umd.edu}).}
\thanks{$\dagger$These authors contributed equally.}%
\thanks{This work is supported by the Office of Naval Research under Grant No. N000141812829.}%
}

\begin{document}
\normalem

\maketitle
\thispagestyle{empty}
\pagestyle{empty}

\begin{abstract}
We propose a risk-aware framework for multi-robot, multi-demand assignment and planning in unknown environments. Our motivation is disaster response and search-and-rescue scenarios where ground vehicles must reach demand locations as soon as possible. We consider a setting  where the terrain information is available only in the form of an aerial, georeferenced image. Deep learning techniques can be used for semantic segmentation of the aerial image to create a cost map for safe ground robot navigation. Such segmentation may still be noisy. Hence, we present a joint planning and perception framework that accounts for the risk introduced due to noisy perception. Our contributions are two-fold: (i) we show how to use Bayesian deep learning techniques to extract risk at the perception level; and (ii) use a risk-theoretical measure, CVaR, for risk-aware planning and assignment. The pipeline is theoretically established, then empirically analyzed through two datasets. We find that accounting for risk at both levels produces quantifiably safer paths and assignments.

\end{abstract}

\section{Introduction}\label{sec:intro}
Consider scenarios where an autonomous ground vehicle must navigate in an unknown environment. Examples include search-and-rescue, space exploration, and disaster response. For instance, consider a disaster response scenario where ground vehicles must supply resources to specific demand locations as soon as possible. In such settings, prior GPS or satellite maps of the environment may no longer be valid. Instead, an aerial robot may be employed to take live aerial images which can then be used to plan the paths of the ground vehicles towards the demand locations. However, due to the inherent uncertainty of aerial images, the paths that are found may not actually represent the situation on the ground. Therefore, there is a risk of the vehicles taking longer to reach the demand positions than planned. In safety-critical situations, one way to mitigate the risk is to assign multiple vehicles to the same demand, with the earliest arriving one actually responding to the demand.


Motivated by this scenario, we study the problem of how to find risk-aware paths for multiple vehicles to serve multiple demand locations. There are two problems to be solved --- assigning the vehicles to the demand locations and finding risk-aware paths from start to demand locations. We present a risk-aware framework to solve both problems simultaneously.

The environment where the vehicles navigate is captured by an overhead image. We implement a deep learning technique for semantic segmentation of the overhead image. Due to the uncertainty from segmentation, the travel cost of the vehicle turns out to be a random variable. \UPDT{Built} on our previous work~\cite{Toubeh2018}, our first contribution is to show how to utilize Bayesian deep  learning techniques to handle the risk from the planning and perception level. After risk-aware planning and perception, we generate a set of candidate paths corresponding to different risk levels from each vehicle's start position to each demand location. Our second contribution is to assign each vehicle to a risk-aware path from its candidate path set to a demand. We utilize a risk measure, Conditional-Value-at-Risk (CVaR), to manage the risk from the uncertainty. Our assignment framework provides the flexibility to trade off between risk and reward, which builds on our previous work~\cite{zhou2018approximation}, with risk here being assessed at multiple levels of the algorithm.
 Our empirical results show that this risk-aware framework results in safer path planning and assignment. 

\paragraph*{Related Work} 
Deep learning has shown significant improvements in perception capabilities for many robotics applications. However, the potential of the positive impact deep learning may have on real-world scenarios is inevitably proportionate to their interpretability and applicability to imperfect environments. In these cases, deep neural networks can even misrepresent data outside the training distribution, giving predictions that are incorrect without providing a clear measure of certainty associated with the result \cite{gal2017phd}. The extraction of uncertainty information, as opposed to the reliance on point estimates, is crucial in safety-critical applications, such as autonomous navigation in an urban, unstructured setting. Methods like Natural-Parameter Networks \cite{wang2016naturalparameter} propose modeling the network parameters and inputs as Gaussian distributions. However, these modifications impose huge computation cost on the model due to the increment in the number of trainable parameters. Lightweight Probabilistic Deep Networks \cite{Gast_2018} alleviates this concern to some extent by making the weights deterministic. Large size networks are also unsuitable for real-time robotics applications which may have constraints on inference time and memory. \cite{gal2016dropout} and \cite{Loquercio_2020} propose methods that allow uncertainty extraction from deep learning models, specifically those that do not interfere with the overall structure or training process. In this work, we leverage \cite{gal2016dropout}, which shows that dropouts, which are often used as a regularization enhancement in neural networks, can provide approximate Bayesian inference and thus help in uncertainty estimation for the deep learning models.   

In addition to considering uncertainty at the perception level, we also utilize some popular risk measures to handle uncertainty at the planning and assignment levels. A typical measure for optimization under uncertainty is the expectation of a stochastic function. However, the expectation is a risk-neutral measure and may not be desirable, especially in critical tasks, like search-and-rescue operations~\cite{majumdar2020should}. For example, we may find a path with a lower expected cost but with high variance. It is quite likely that the vehicle may encounter a much larger cost (as compared to the expected one) when traveling on this path. Thus, instead of using expected cost, we utilize some other risk-aware measures, such as mean-variance~\cite{marcus1997risk,chung2019risk} and conditional-value-at-risk (CVaR)~\cite{rockafellar2000optimization,majumdar2020should}. 

In particular, we use mean-variance as the \UPDT{\sout{risk/cost} risk-aware cost} measure in the A* planner~\cite{lavalle1998rapidly} for planning paths for the vehicles. The mean-variance measure allows us to balance the mean cost and uncertainty (variance) when planning paths. Also, for the path assignment, we use CVaR to deal with the uncertainty on the path level. $\text{CVaR}_{\alpha}$ explicitly takes into account the risk associated with bad scenarios~\cite{rockafellar2000optimization, majumdar2020should}. Specifically, $\text{CVaR}_{\alpha}$ measures the expectation of a random variable in the $100\alpha$--percentile worst scenarios. Here, the user-defined risk-level, $0<\alpha\leq 1$, provides a user with the flexibility to choose a risk that they would like to take. Setting $\alpha=1$ makes $\text{CVaR}_{\alpha}$ equal to the expectation whereas $\text{CVaR}_{\alpha}\approx 0$ is akin to worst-case optimization.

\UPDT{Risk in autonomous systems ultimately involves the risk of failure, whether due to damage to the agents or their environment or mission failure. Several risk-aware planning frameworks have been proposed for collision avoidance, such as the dynamic risk density function in ~\cite{pierson2019dynamic} which infers congestion cost from sensor inputs. Our navigation cost function observes the risk associated with noisy sensor perception and do not assume sensor perception is always correct. Uncertain semantic maps have also been proposed for perception-based planning. Both ~\cite{kantaros2019optimal} and ~\cite{fu2016optimal} model an unknown environment as a semantic map by assuming Gaussian distributions over landmark positions for simultaneous localization and planning (SLAM). These address the uncertainty in planning, but not in perception, i.e. uncertain landmark classification.}

\textbf{Contributions.} In this paper, we have three main contributions. 
\begin{itemize}
    \item We present a framework that plans and assigns risk-aware paths for robots that navigate in unknown environments. 
    \item We utilize the Bayesian deep learning technique to learn an unknown environment whose information is only available by an overhead, georeferenced image. 
    \item We deal with the uncertainty at both path planning and assignment levels by optimizing the corresponding risk measures. In the end, we assign each vehicle a risk-aware path to a demand location and the path assignment is guaranteed to have a bounded approximation performance of the optimal \UPDT{assignment}.
\end{itemize}
This work builds on our prior work where we studied these two problems (uncertainty extraction from deep learning and CVaR optimization) individually. Here, we investigate the joint problem. We find that utilizing the uncertainty extraction from deep learning and managing the risk from uncertainty by CVaR optimization provide the vehicles with safer and risk-aware paths in unknown environments.

\section{Preliminaries}\label{sec:background}
We start by defining the notations used in the paper. We give background on a risk measure, conditional-value-at-risk (CVaR). \UPDT{We then provide a formal definition of the joint problem of planning and assignment.}

\subsection{Conditional Value at Risk}~\label{subsec:CVaR}
Let $X$ be a random variable. $\text{CVaR}_{\alpha}(X)$ denotes the expectation on the $\alpha$-worst scenarios of \UPDT{the utility or cost function} $f$ with $\alpha \in (0,1]$. More specifically, if $X$ indicates reward or benefit,  $\text{CVaR}_{\alpha}(X)$ denotes the expectation on the left $\alpha$-tailed scenarios. While, if $X$ represents loss or penalty, $\text{CVaR}_{\alpha}(X)$ is the expectation on the right $\alpha$-tailed cases. Here, $\alpha$ is the confidence level or the risk level. If $\alpha$ is close to $0$, $\text{CVaR}_{\alpha}$ is close to the worst-case. If $\alpha$ is equal to $1$, $\text{CVaR}_{\alpha}$ is same as the expectation. 

\begin{figure}[h]
  \centering
\includegraphics[width=3in]{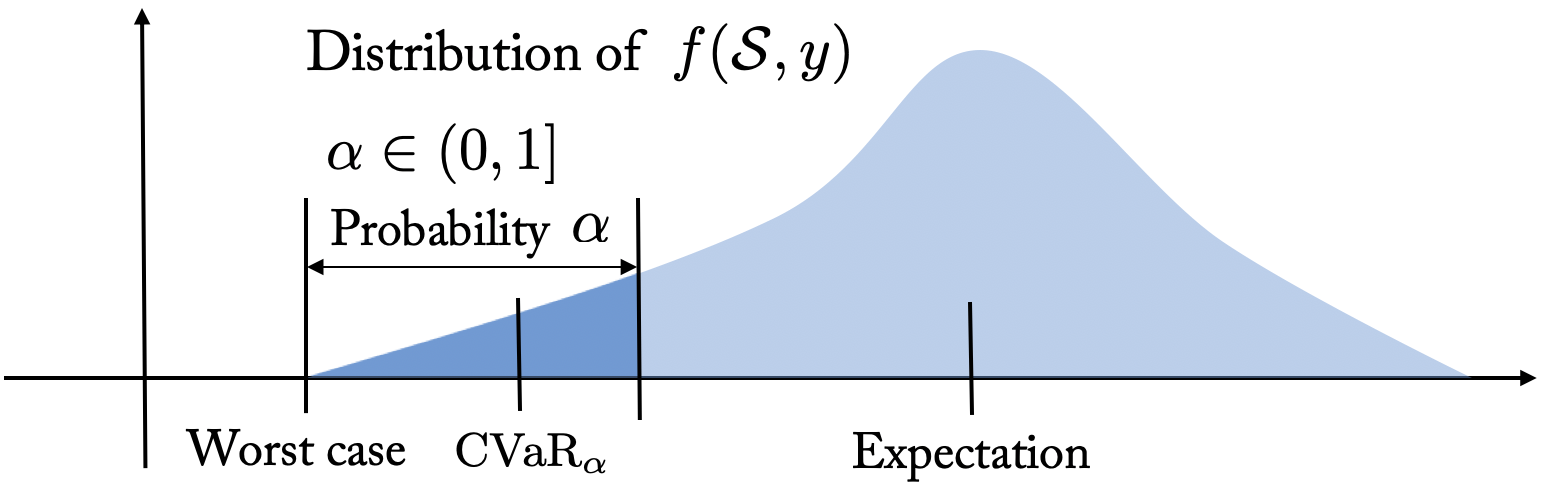}
  \caption{$\text{CVaR}_{\alpha}$ of function $f(\mathcal{S},y)$.\label{fig:cvar}}
\end{figure}

In this paper, the utility function $f(\mathcal{S},y)$ defined on set  $\mathcal{S}$ is a random variable with randomness induced by parameter $y$. Since $f(\mathcal{S},y)$ is utility that indicates benefit, $\text{CVaR}_{\alpha}[f(\mathcal{S},y)]$  denotes the expectation on the left $\alpha$-tailed cases, as shown in  Figure~\ref{fig:cvar}. 

We generally maximize $\text{CVaR}_{\alpha}[f(\mathcal{S},y)]$ by solving 
\begin{equation}
  \max_{\mathcal{S}, \tau} ~\tau - \frac{1}{\alpha}\mathbb{E}[(\tau-f(\mathcal{S},y))_{+}],
  \label{eqn:cvar_auxiliary}
\end{equation}
where $\mathcal{S}$ is a decision set (or solution set), \UPDT{$(t)_+ = \max(t,0)$}, and $\tau\in \mathbb{R}_+$ is an auxiliary parameter. For the ease of expression, we define 
\begin{equation}
    H(\mathcal{S},\tau) = \tau - \frac{1}{\alpha}\mathbb{E}[(\tau-f(\mathcal{S},y))_{+}]
    \label{eqn:aux_fun}
\end{equation}

\subsection{Problem Formulation}~\label{sec:prob_form}
We consider the problem of finding paths for multiple vehicles to serve multiple demand locations \UPDT{\sout{(Figure~\ref{figs:MoD})}}. In particular, we are given $N$ vehicles' start positions, $\mathcal{V}=\{1,\cdots, N\}$ and $M$ demand locations, $\mathcal{D}=\{1, \cdots, M\}$ in the environment. The environment is represented by an overhead, georeferenced RGB image as shown in Figure~\ref{fig:overview}. The goal is to find offline paths for each vehicle such that they collectively serve all the demands using navigation costs derived from the overhead images. 



The cost of a path in the environment can be estimated by first performing a semantic segmentation of the overhead image. However, semantic segmentation is typically imperfect~\cite{gal2017phd}, and as such the estimated cost of a path may not be accurate. The problem we address in this paper is that of finding paths for vehicles to collectively serve all demands under travel-cost uncertainty.  

We are motivated by tasks that are urgent and time-critical, such as fighting fires~\cite{harikumar2018multi} and delivering medical supplies in emergencies~\cite{ackerman2018medical}. When the number of vehicles is more than the demands, assigning multiple redundant vehicles to demands helps counter the effect of uncertainty~\cite{prorok2019redundant}. The waiting time at a demand location is the time taken by the earliest vehicle to arrive at that location. If the travel times are deterministic, then it is known in advance which vehicle will arrive first. When travel times are uncertain, as in this work, the arrival time of the earliest vehicle itself is a random variable. The goal is to assign vehicles to demand locations and find corresponding paths for the vehicles from the start to the assigned demand locations.

For convenience, we convert the minimization problem into a maximization one \UPDT{to make it submodular} by taking the reciprocal of the travel cost. Specifically, we use the travel \textit{efficiency}, the reciprocal of travel cost, as the measure. Thus, the travel efficiency of a demand location is the maximum of the travel efficiencies of the vehicles that reach this demand location. 
The overall travel efficiency, denoted by $f$ is the sum of the travel efficiencies of all demand locations. Notably, $f$ is also a random variable. 

Our goal is to find risk-aware paths from vehicles' start positions to demand locations given a user-defined risk level $\alpha$. We formulate a risk-aware path finding problem by maximizing $\text{CVaR}_{\alpha}$ on the travel efficiency (Problem~\ref{prob:risk_plan_find}).  
\begin{problem}[Risk-Aware Path Finding]~\label{prob:risk_plan_find}
\begin{equation}
\underset{\mathcal{S}\subseteq \mathcal{X}}{\max}~ \emph{CVaR}_{\alpha}[f(\mathcal{S},y)]
\label{eqn:cvar_max}
\end{equation}
\label{pro:cvar_max}
\end{problem}
where $\mathcal{S}$ is a set of paths for vehicles \UPDT{\sout{with}} \UPDT{(}``per path per vehicle''\UPDT{)}, $\mathcal{X}$ is a ground set of paths from which $\mathcal{S}$ is chosen, and $f(\mathcal{S}, y)$ is the travel efficiency on the path set $\mathcal{S}$, with randomness induced by $y$.  




\section{Framework and Analysis}~\label{sec:alg_ana}
\begin{figure*}[h]
  \centering
  \includegraphics[width=1.8\columnwidth]{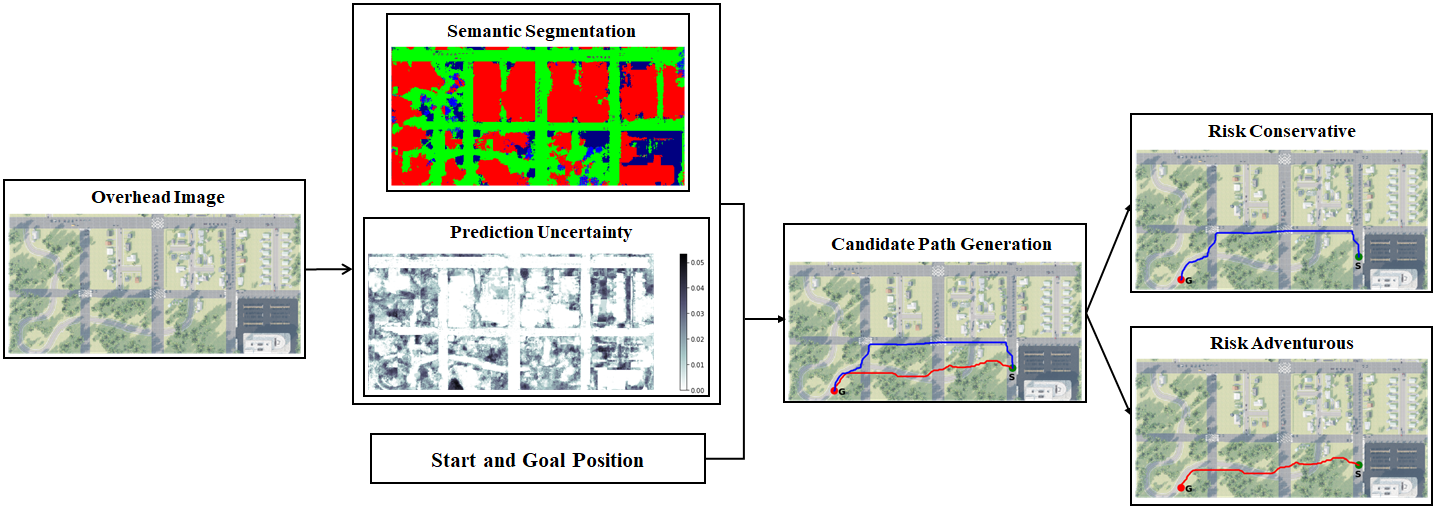}
  \caption{The breakdown of the framework's parts. Given an overhead image input, the algorithm provides a semantic segmentation and uncertainty map, then generates candidate paths, and finally performs the risk-aware path assignment of vehicles to demands.}
  \label{fig:overview}
\end{figure*}
The framework we propose consists of three main parts: semantic segmentation, candidate path generation, and risk-aware assignment. The overall pipeline of the framework and its parts are shown in Figure~\ref{fig:overview}. The inputs to the framework are a single overhead aerial two-dimensional image, the vehicles' start positions, and the demand locations. The output is a risk-aware assignment of paths from vehicles' start positions to demand locations. 

The input is first semantically segmented into per pixel labels. These labels are assigned a cost proportionate to the risk involved in traversing them. The cost map and the uncertainty associated with the segmentation are then used as inputs to a path planner which generates candidate paths for assignment. Finally, the candidate paths from each vehicle's start position to each demand location are computed by maximizing CVaR, for risk-aware path assignment.
\subsection{Semantic Segmentation}
To plan a path for a vehicle in the environment, we need to first recognize features in the image, such as road, person, car, and so on. Given an input aerial image provided by an unmanned aerial vehicle (UAV), a deep learning model is used to provide a semantically segmented map. We then utilize an approximation technique, using dropout, to learn the uncertainty in the semantic labels provided by the model~\cite{gal2016dropout}. The advantage of using uncertainty has been studied~\cite{gal2017phd}. After semantic segmentation and uncertainty extraction, we assign different cost values for the robot traveling on different terrain features\UPDT{, similar to the cost function in~\cite{Toubeh2018}}. For example, the cost of traveling on the road is less than that of traveling on vegetation. Traveling on a person and cars is impossible, and thus, the corresponding cost can be set to infinity. Notably, the cost can be the energy or time spent by a vehicle traveling on the terrains. Since there exists uncertainty in the terrain features provided by the deep learning model, the cost of traveling on them is also a random variable with some uncertainty.

\subsection{Candidate Path Generation}~\label{subsec:path_gen}
Once a risk-aware cost map is generated based on the semantic segmentation of the terrain, augmented by the confidence in prediction, candidate paths are generated from each vehicle location to each demand location by \UPDT{an} A* planner. For the different combinations of vehicles and demands, a candidate path represents a potential feasible route given on the \UPDT{\sout{aerial}} map. Relying on deep learning segmentation alone can be risky. Given confidence information, we expect paths to avoid regions of high uncertainty which could involve out-of-distribution data or misclassifications, as shown in Figure~\ref{fig:segnet_out1}. 


For \UPDT{the} A* planner, a risk-aware cost function $C(x)$ is defined on each pixel $x$. It combines the cost associated with the classified terrain type and the variance.
\UPDT{\sout{
\begin{equation}
 C(x)  :=  C(l_x) + \lambda \texttt{Var}(x),
 \label{eqn:cost_assign}
\end{equation}
}}

\begin{equation}
 \hat{C}(x)  :=  C(l_x) + \lambda \texttt{Var}(l_x),
 \label{eqn:cost_assign}
\end{equation}
where $C$ refers to the risk-neutral mean cost associated with the most likely predicted label $l_{x}$ for a pixel $x$ given multiple softmax outputs. The risk-aware function $\hat{C}$ assigns a cost to each pixel $x$, characterized by a user-defined cost mapping and a weight parameter $\lambda$ to quantify emphasis on the uncertainty. We use the variance in the prediction as a measure of uncertainty.
\subsection{Risk-Aware Path Assignment}~\label{subsec:path_assign}
Consider a set of $K$ candidate paths $\mathcal{P}=\{1,\cdots, K\}$ \UPDT{\sout{\UPDS{that} are}} generated from each vehicle's start position to each demand location in Section~\ref{subsec:path_gen}. We then assign each vehicle a path to a demand location. As mentioned in Section~\ref{sec:prob_form}, we follow a redundant assignment setting~\cite{prorok2019redundant}, where each vehicle can be assigned to at most one demand location and multiple vehicles can be assigned to the same demand location. \UPDT{\sout{Only the vehicle through a path arrives with the highest efficiency is chosen at each demand location.}} \UPDT{Only the vehicle that arrives through a path with with the highest efficiency is chosen at each demand location.}

\UPDT{\sout{Denote}} The travel efficiency \UPDT{is denoted} for a vehicle starting at  $i\in \mathcal{V}$ arriving at demand location $j\in\mathcal{D}$ through path $k \in \mathcal{P}$ as $e_{ijk}$. Correspondingly, we take the tuple $(i,j,k)$ as an assignment where the vehicle-path pair $(i,k)$ is assigned to the demand location $j$. \UPDT{\sout{Denote}} The total efficiency \UPDT{is denoted} at all $M$ demand locations as 
\begin{equation}
 f(\mathcal{S}, y)  = \sum_{j\in M} \max_{(i,j,k)\in \mathcal{S}_j} ~e_{ijk} 
 \label{eqn:fsy_assign}
\end{equation}
where $\mathcal{S}_j$ denotes the assignment set $\{(i,j,k)\}$ to the demand location $j$. The total assignment set $\mathcal{S}:= \bigcup_{j=1}^{M} \mathcal{S}_j$ is a collection of the assignment sets at all demand locations. Notably, since each vehicle-path pair $(i,k)$ can be assigned to at most one demand location, all $\mathcal{S}_j$(s) are disjoint, i.e., $\mathcal{S}_j \cap \mathcal{S}_{j'} = \emptyset, ~\forall j\ne j',  j, j'\in\mathcal{D}$. This is called a partition matroid constraint in the literature~\cite{fisher1978analysis}, \UPDT{\sout{which we denote}}\UPDT{denoted} by $\mathcal{I}$. 

We use the ``max" operator to capture the selection rule that only the vehicle-\UPDT{path }pair $(i,k)$ with the maximum efficiency is chosen at each demand location. Due to the ``max" operator, the total efficiency $f(\mathcal{S},y)$ turns out to be monotone submodular in $\mathcal{S}$. If there is no assignment, we set $f(\emptyset, y)=0$ to \UPDT{\sout{make $f$ normalized}}\UPDT{normalize $f$}. Here, $y$ indicates the randomness of $f(\mathcal{S},y)$ due to the uncertainty in travel efficiency. 

Then, by utilizing Equation~\ref{eqn:cvar_auxiliary} and the candidate paths generated in Section~\ref{subsec:path_gen}, we transform Problem~\ref{prob:risk_plan_find} to a risk-aware path assignment problem below.  
\begin{problem}[Risk-Aware Path Assignment]~\label{prob:risk_plan_assign}
\begin{align}
\underset{\mathcal{S}\subseteq \mathcal{X}}{\max} ~~\tau - \frac{1}{\alpha}\mathbb{E}[(\tau-\sum_{j\in M} \max_{(i,j,k)\in \mathcal{S}_j} ~e_{ijk})_{+}]\nonumber\\
s.t.~\mathcal{S}= \bigcup_{j=1}^{M} \mathcal{S}_j, ~\mathcal{S} \in \mathcal{I} ~~~and 
~~~~\tau\in[0, \Gamma],
\label{eqn:cvar_max}
\end{align}
\label{pro:cvar_max}
\end{problem}
with $\mathcal{S}$ the path assignment set (``per path per vehicle''), $\mathcal{I}$ the partition matroid constraint~\cite{fisher1978analysis}, $\mathcal{X}$ the ground set of all possible assignments, $\{(i,j,k)\},~k\in\mathcal{P},~i\in\mathcal{V},~j\in\mathcal{D}$,  and $\Gamma\in \mathbb{R}^{+}$ the upper bound of the parameter $\tau$. $\Gamma$ can be set as an upper bound on $f(\mathcal{S}, y)$ (Eq.~\ref{eqn:fsy_assign}).

Building on the sequential greedy algorithm (\texttt{SGA}) from our previous work \cite{zhou2018approximation}, we present Algorithm~\ref{alg:sga} for solving Problem~\ref{prob:risk_plan_assign}. 

\begin{algorithm}[t]
\caption{Risk-Aware Path Assignment}  
\begin{algorithmic}[1]
\REQUIRE 
\begin{itemize}
\item Vehicles' initial positions $\mathcal{V}$, demand\\ locations $\mathcal{D}$, and path set $\mathcal{P}$
\item User-defined risk level $\alpha \in [0, 1]$
\item Range of the parameter $\tau \in [0, \Gamma]$ and discretization stage $\Delta \in (0, \Gamma]$
\item An oracle $\mathcal{O}$ that approximates $H(\mathcal{S}, \tau)$ \\as $\hat{H}(\mathcal{S}, \tau)$
\end{itemize}
\ENSURE 
\begin{itemize}
\item Path assignment $\mathcal{S}$
\end{itemize}

\STATE $\mathcal{M}\leftarrow\emptyset$ \label{line:initiliaze}
\FOR{$ ~i \in \{0,1,\cdots, \ceil{\frac{\Gamma}{\Delta}}\}$}
\label{line:search_tau_forstart}
\STATE $\tau^i = i\Delta$\label{line:search_tau_separation}
\STATE $\mathcal{S}^{i}\leftarrow\emptyset$ \label{line:gre_empty}
\FOR{$l=1:|\mathcal{D}|$}
\label{line:D_round}
\STATE $(i^\star, j^\star, k^\star) = \underset{i\in \mathcal{V}, j \in \mathcal{D}, k \in \mathcal{P}}{\text{argmax}}~\hat{H}(\mathcal{S}^{i}\cup (i,j,k), \tau^i) - \hat{H}(\mathcal{S}^{i}, \tau^i)$ \label{line:max_margin}
\STATE $\mathcal{S}^{i}\leftarrow \mathcal{S}^{i} \cup (i^\star,j^\star,k^\star)$
\label{line:add_in_set}
\STATE $\mathcal{V} = \mathcal{V}\setminus i^\star$
\label{line:delete_vstar}
\ENDFOR
\label{line:greedy_end}
\STATE $\mathcal{M} = \mathcal{M} \cup \{(\mathcal{S}^{i}, \tau^i)\}$\label{line:pair_collection}
\ENDFOR\label{line:search_tau_forend}
\STATE $(\mathcal{S}, \tau^{\star}) = \underset{(\mathcal{S}^{i}, \tau^i) \in \mathcal{M}}{\text{argmax}}~{\hat{H}(\mathcal{S}^{i}, \tau^i)}$ \label{line:find_best_pair}
\end{algorithmic}
\label{alg:sga}
\end{algorithm}

There are four stages in Algorithm~\ref{alg:sga}: 
\paragraph{Initialization (line~\ref{line:initiliaze})} We initialize a storage set $\mathcal{M}$ to be empty. We use $\mathcal{M}$ to store the assignment $\mathcal{S}$ with the corresponding $\tau$ when searching all possible values of $\tau$. 

\paragraph{Searching for $\tau$ (\textbf{for} loop in lines~\ref{line:search_tau_forstart}--\ref{line:search_tau_forend})} We sequentially search for all possible values of $\tau$ within $[0, \Gamma]$ by a user-defined separation $\Delta$ (line~\ref{line:search_tau_separation}). 
$\Gamma$ is an upper bound on $\tau$ and can be set \UPDT{\sout{by the user }}based on the specific problem at hand. We show how to compute $\Gamma$ in specific scenarios in Section~\ref{sec:sim}. 

\paragraph{Greedy algorithm (lines~\ref{line:gre_empty}--\ref{line:greedy_end})} For a specific $\tau$, e.g., $\tau_i$, we use the greedy approach~\cite{fisher1978analysis} to choose the corresponding assignment set $\mathcal{S}^{i}$. We first initialize set $\mathcal{S}^{i}$ to be the empty set (line~\ref{line:gre_empty}). Then we execute the greedy algorithm in $|\mathcal{D}|$ rounds (line~\ref{line:D_round}), since the total number of demand locations to be served is $\mathcal{D}$. At each round, the assignment $(i^\star,j^\star,k^\star)$ which gives the maximum marginal gain of $\hat{H}(\mathcal{S}^{i}, \tau^i)$ is selected (line~\ref{line:max_margin}) and added into set $\mathcal{S}^{i}$ (line~\ref{line:add_in_set}). Then, we remove the vehicle position $i^\star$ from  $\mathcal{V}$ (line~\ref{line:delete_vstar}) to make sure vehicle position $i^\star$ and the corresponding paths will never be selected in the following rounds. 

\paragraph{Find\UPDT{ing} the best assignment (line~\ref{line:find_best_pair})} From the collection of $(\mathcal{S},\tau)$ pairs, $\mathcal{M}$ (line~\ref{line:pair_collection}), we pick the one that maximizes $\hat{H}(\mathcal{S}^{i}, \tau^i)$ as the output $\mathcal{S}$ with the corresponding $\tau$ denoted by $\tau^\star$ (line~\ref{line:find_best_pair}). 

\textit{Oracle Design:} We use an oracle $\mathcal{O}$ to calculate the value of $H(\mathcal{S}, \tau)$. The oracle uses a sampling based method to approximate $H(\mathcal{S}, \tau)$~\cite{zhou2018approximation}. 

It has been shown in \cite{zhou2018approximation} that Algorithm~\ref{alg:sga} generates an assignment that has the bounded approximation performance of the optimal assignment. 

\section{Evaluation}~\label{sec:sim}
In this section, we report the results from empirical studies evaluating the proposed risk-aware perception\UPDT{\sout{and}, }planning\UPDT{, and assignment} framework. We start by describing the experimental setup and then describe the results.

\textbf{Setup.} We use \UPDT{the} AirSim~\cite{shah2018airsim} simulator as it offers photo-realistic images along with the semantically segmented ground truth in \UPDT{\sout{multiple}}\UPDT{various} pre-defined environments. We use \UPDT{the} \textit{CityEnviron} environment which contains city-like and suburban landscapes. We collect the down\UPDT{ward}-looking aerial images at an altitude value of 200m in the simulator. The dataset thus generated has 480 images, which are divided in the ratio of 10:3:3 as training, validation, and testing data for the Bayesian SegNet~\cite{kendall2015bayesian}. We use the PyTorch implementation of a basic version of this network~\cite{mshahsemseg} and add dropout layers in accordance with the original architecture. 
In our ground truth, we reduce the ground truth to 12 classes consistent with the original model~\cite{kendall2015bayesian}. 

\begin{figure}[t]
\centering
\subfigure[Output from training on shuffled data]{
  \includegraphics[clip,width=0.9\columnwidth]{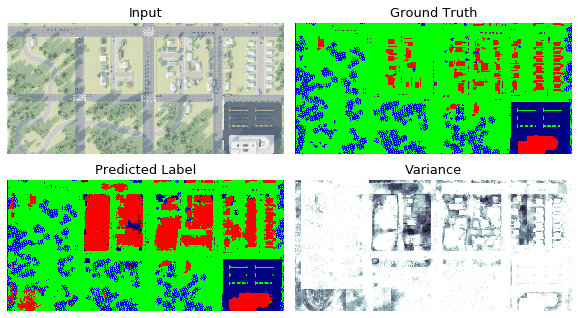}
}

\subfigure[Output from training on unshuffled data]{
  \includegraphics[clip,width=0.9\columnwidth]{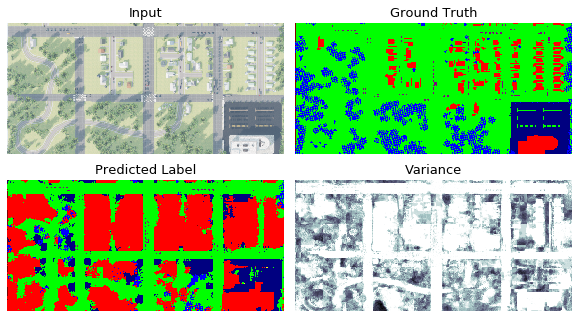}
}
\caption{Difference in variance due to data distribution. \label{fig:segnet_out1}}

\end{figure}

Due to limitations in the simulator, roads and grass patches are indistinguishable in the ground truth as depicted in Figure~\ref{fig:segnet_out1}. We keep such images limited to the test dataset. Hence, the grass patches act as unknown objects (similar to out-of-distribution) to the model and provide interesting observations on the uncertainty in prediction for such objects.


\UPDT{\sout{We produce 20 outputs (or \textit{stochastic samples}) for each image using the trained Bayesian SegNet. For each pixel, the predicted label \UPDT{is} found as the most frequent label among all \UPDT{of} the most likely labels for each pixel, i.e., $l_{p,ml} = \texttt{Mode}_\text{samples}(\text{arg} \max\limits_c P(c|x))$ for a pixel $p$, 
where $x$ is the input to the model, $c$ is the class/label, and $\texttt{Mode}$ stands for the statistical mode. Uncertainty for each pixel is defined as the average variance in the probabilities of labels i.e. $\texttt{uncertainty}(p) = \frac{1}{c} \sum_c \texttt{Var}( P(c|x) )$. }} \UPDT{We produce 20 outputs (or \textit{stochastic samples}) for each image using the trained Bayesian SegNet. For each pixel, the predicted label is found as the most frequent label among all of the most likely labels for each pixel, i.e., $l_{x} = \texttt{Mode}_\text{samples}(\text{arg} \max\limits_c P(c|x))$ where $x$ is the pixel input to the model, $c$ is the class label, and $\texttt{Mode}$ stands for the statistical mode. Uncertainty for each pixel is defined as the average variance in the probabilities of labels i.e. $\texttt{uncertainty}(x) = \frac{1}{c} \sum_c \texttt{Var}( P(c|x) )$. }The cost for each pixel is calculated using these two quantities, which is further passed to the path planning algorithm (Sec.~\ref{subsec:path_gen}). To provide an orthographic view in the results, we make predictions over partially overlapping images with perspective projections. Then we combine subsets around the center of the images.  

\textbf{Results.} 
\textit{\UPDT{a) Out-of-Distribution Data.}}
The Bayesian SegNet model used in our path planning algorithm has dissimilar data distribution for training and testing. The effect of this is shown in Figure~\ref{fig:segnet_out1}. When the data distributions are the same across training and test dataset (shuffled data), the model is able to make predictions with very low variance (implied by brighter shades). However, in the case of dissimilar distribution, some part of the image acts as unseen data and thus the model prediction has high variance. This highlights the importance of uncertainty. For example, the prediction is precise for already observed data and thus \UPDT{the} decision involving such input would be less risky\UPDT{, whereas the opposite is true for unobserved data}.

\begin{figure}
  \centering
  \includegraphics[width=0.9\columnwidth]{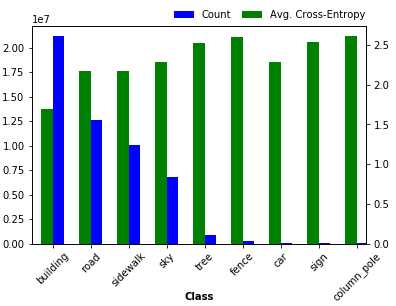}
  \caption{Average Cross-Entropy for each class/label.}
  \label{figs:avg_cross_ent}
\end{figure}

\textit{\UPDT{b) Average Cross-Entropy on the Test Dataset.}}
In order to understand the correlation between the quality of prediction and the training data, we look at the cross-entropy of the model prediction (over unshuffled data) averaged over the 20 samples and the number of pixels in the training data for each class in Figure~\ref{figs:avg_cross_ent}. 
The model performs comparatively well in identifying obstacles like buildings. Objects which are rarely observed in the ground truth have very high cross-entropy. The lack of difference between the \UPDT{\sout{ground} vegetation} and the road in the ground truth also affects the performance of relevant classes. As expected, high cross-entropy is observed for classes where the pixel count is low. 


\textit{\UPDT{c) Risk-Aware Path Planning and Assignment.}}

\begin{figure}
  \centering
  \includegraphics[width=0.7\columnwidth]{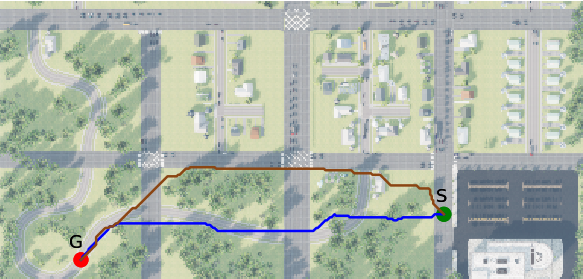}
  \caption{Effect of $\lambda$ on path planning. {\color{blue} Smaller $\lambda$ (=10)}  gives a shorter path with high uncertainty; {\color{brown} larger $\lambda$ (=50)} gives a longer path with low uncertainty.}
  \label{fig:path_lambda}
\end{figure}

\UPDT{The} value of $\lambda$ decides the risk-awareness of path planning. This effect is shown in Figure~\ref{fig:path_lambda} where the cost  \UPDT{\sout{of each navigable class is 1 and for non-navigable classes, the cost is 3 (except for the tree class, where it is 2)}}\UPDT{is 1 for the navigable classes and 3 for non-navigable classes (except for the tree class, where it is 2)}. For a small value of $\lambda$, the algorithm plans a short path passing through the vegetation. For a high value of $\lambda$, this area is avoided due to \UPDT{the} high uncertainty in the prediction for this part of the scene. 




By varying the values of $\lambda$, we generate $K=2$ candidate paths \UPDT{\sout{from each vehicle's start position to each demand location}}\UPDT{for each pair of start and demand location}. We consider assigning $N=3$ supply vehicles to $M=2$ demand locations in this unknown environment \UPDT{\sout{(Fig.~\ref{fig:path_lambda})}}.

\begin{figure}[t]
\centering{
\subfigure[Demand location $d_1$]
{\includegraphics[width=0.48\columnwidth]{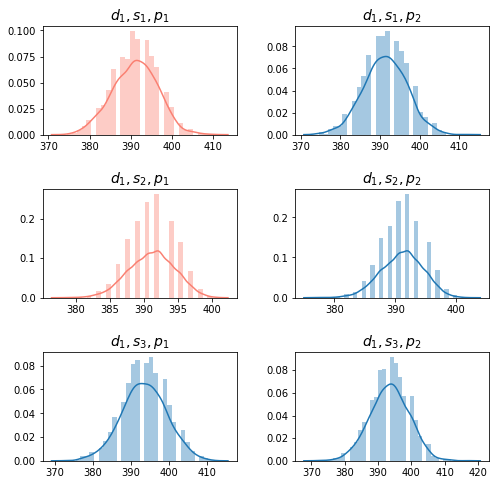}}~~~
\subfigure[Demand location $d_2$]
{\includegraphics[width=0.48\columnwidth]{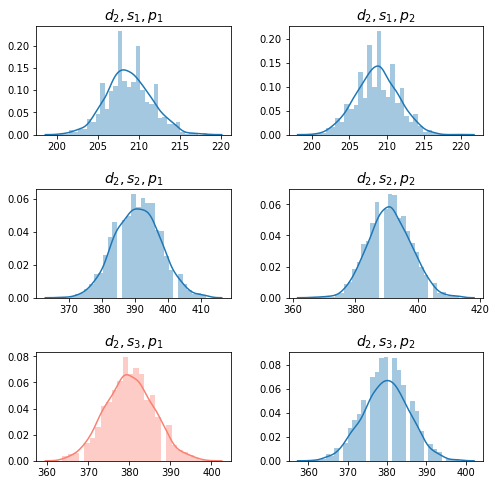}}
\caption{Efficiency distributions of paths and the path assignment when $\alpha = 0.01$. The assigned path for each robot is marked in red. \label{fig:risk_001}}
}
\end{figure}

\begin{figure}[t]
\centering{
\subfigure[Demand location $d_1$]
{\includegraphics[width=0.48\columnwidth]{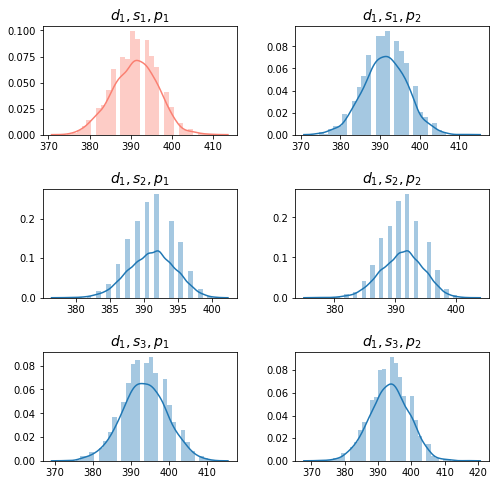}}~~~
\subfigure[Demand location $d_2$]
{\includegraphics[width=0.48\columnwidth]{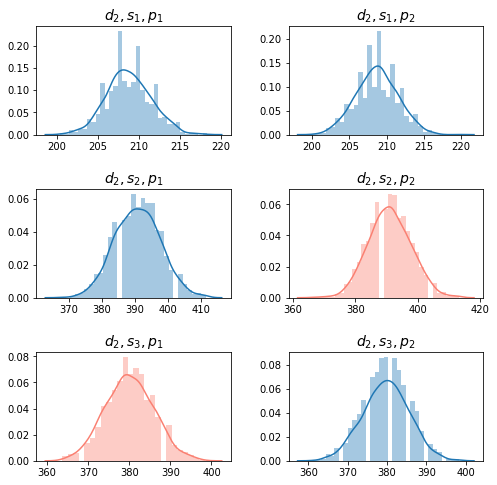}}
\caption{Efficiency distributions of paths and the path assignment when $\alpha = 1$. The assigned path for each robot is marked in red.  \label{fig:risk_1}}
}
\end{figure}

Due to the \UPDT{\sout{imperfectness of} imperfect} semantic segmentation, the \UPDT{\sout{cost/efficiency} cost or efficiency} of the path is a random variable. We show the efficiency distributions of the paths from \UPDT{the} vehicles' start positions to \UPDT{the} demand locations in Figure~\ref{fig:risk_001} and Figure~\ref{fig:risk_1}. 

We use Algorithm~\ref{alg:sga} to assign each vehicle a risk-aware path to a demand location. 
For example, in Figure~\ref{fig:risk_001}-(a), vehicle $v_2$ is assigned path $p_1$ for demand location $d_1$ when the risk level is small, e.g., $\alpha = 0.01$. In contrast, \UPDT{as seen in Figure~\ref{fig:risk_1}-(a)}, when the risk level is high, e.g., $\alpha=1$, the assignment is more adventurous, and thus the path with a larger mean efficiency and a larger variance is selected. As shown in Figure~\ref{fig:risk_1}-(b), vehicle $v_2$ switches to path $p_2$ for demand location $d_2$, because the efficiency of this path has a larger mean. The path assignment changes follow the comparison of CVaR values, i.e., $\text{CVaR}_{0.01}[e(p_1)] > \text{CVaR}_{0.01}[e(p_2)]$ and $\text{CVaR}_{1}[e(p_1)] < \text{CVaR}_{1}[e(p_2)]$ with $e(p_1)$ and $e(p_2)$ denoting the efficiency of path $p_1$ and $p_2$, respectively. Thus, the risk level, $\alpha$, provides\UPDT{\sout{a user with}} the flexibility to trade off between risk and total efficiency (reward). 

\begin{figure}[t]
  \centering
  \includegraphics[width=0.8\columnwidth]{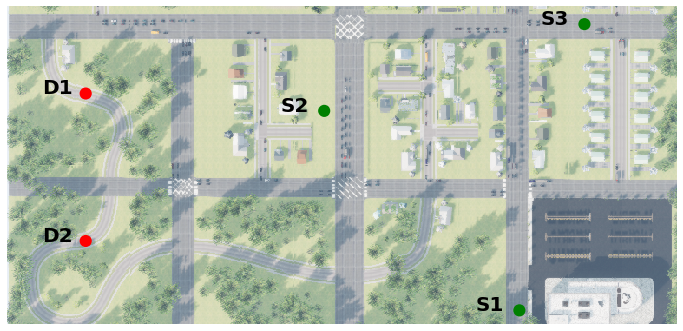}
  \caption{Start and demand positions for \textit{surprise} calculations.}
  \label{fig:city_start_goal}
\end{figure}

\begin{figure}[t]
  \centering
  \includegraphics[width=0.65\columnwidth]{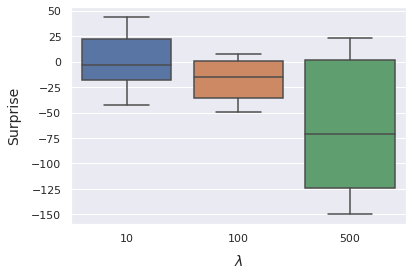}
  \caption{\textit{Surprise} vs $\lambda$.}
  \label{fig:all_surprise}
\end{figure}

\textit{\UPDT{d) Quantitative results.}}
In order to quantify the effect of $\lambda$, we define a metric called \textit{surprise}  for a path as the difference  between the cost of the ground truth labels and predicted labels. We take 6 combinations of start and demand positions given in Figure~\ref{fig:city_start_goal} 
and find the average value of \textit{surprise} as shown in Figure~\ref{fig:all_surprise}. 
Ideally, we expect the \textit{surprise} to reduce with emphasis on variance. However, for large values of $\lambda$, even a few pixels with high variance may greatly increase the cost. In general, a higher value of $\lambda$ may cause the robot to choose a longer path, which may increase the cost of traversal. However, sometimes the path may pass though only navigable regions, resulting in a small value of  \textit{surprise}. This causes the \textit{surprise} to have a larger variance. Thus, the value of $\lambda$ should be chosen after careful consideration of the variance in \UPDT{\sout{model prediction}}\UPDT{predictions} and the range of the cost mapping. 

\begin{figure}[t]
  \centering
  \includegraphics[width=0.9\columnwidth]{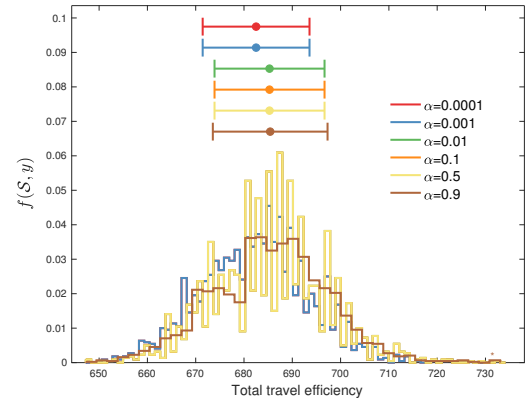}
  \caption{Distribution of the total travel efficiency $f(\mathcal{S}, y)$ by Algorithm~\ref{alg:sga}.}
  \label{fig:dis_total_effi}
\end{figure}

We also plot the distribution of \UPDT{the} total travel efficiency (Eq.~\ref{eqn:fsy_assign}) in Figure~\ref{fig:dis_total_effi}.
$\mathcal{S}$ is the path assignment selected by Algorithm~\ref{alg:sga}. With small risk levels $\alpha$, paths with low efficiencies (equivalently, low uncertainty) are mostly selected. This is because a small risk level indicates the assignment is conservative and only willing to take a \UPDT{\sout{little} small amount of} risk. Thus, the path with lower efficiency and lower uncertainty is assigned to avoid the risk induced by the uncertainty. In contrast, when $\alpha$ is large, the assignment is riskier. In such a case, the paths with high efficiencies (equivalently, high uncertainty) are selected.

\section{Conclusion} \label{sec:conslude}
In this paper, we propose a risk-aware path planning and assignment framework for vehicles navigating in an unknown environment. We consider a scenario \UPDT{\sout{that} in which} the information of the unknown environment is only available by an overhead, georeferenced image that is taken by an aerial robot. To deal with this challenge, we utilize Bayesian deep learning to learn the environment though semantic segmentation. Since the output of this segmentation is noisy, the cost of the vehicle traversing in the environment is uncertain. To deal with the cost uncertainty, we optimize some popular risk measures to generate and assign risk-aware paths for the vehicles. We use extensive simulation results to show the effectiveness of our risk-aware strategy.

Future work will focus on the online coordination of aerial and ground vehicles to achieve long-term, real-time risk-aware navigation in unknown environments.  

\bibliographystyle{IEEEtran}
\bibliography{refs.bib}

\end{document}